# Towards Instance Segmentation with Object Priority: Prominent Object Detection and Recognition

Hamed R. Tavakoli
Department of Computer Science
Aalto University, Espoo, Finland
hamed.r-tavakoli@aalto.fi

Jorma Laaksonen
Department of Computer Science
Aalto University, Espoo, Finland
jorma.laaksonen@aalto.fi

## Abstract

*This manuscript introduces the problem of* prominent object detection and recognition *inspired by the fact that human seems to priorities perception of scene elements. The problem deals with finding the most important region of interest, segmenting the relevant item/object in that area, and assigning it an object class label. In other words, we are solving the three problems of saliency modeling, saliency detection, and object recognition under one umbrella. The motivation behind such a problem formulation is (1) the benefits to the knowledge representation-based vision pipelines, and (2) the potential improvements in emulating bio-inspired vision systems by solving these three problems together. We are foreseeing extending this problem formulation to fully semantically segmented scenes with instance object priority for high-level inferences in various applications including assistive vision. Along with a new problem definition, we also propose a method to achieve such a task. The proposed model predicts the most important area in the image, segments the associated objects, and labels them. The proposed problem and method are evaluated against human fixations, annotated segmentation masks, and object class categories. We define a chance level for each of the evaluation criterion to compare the proposed algorithm with. Despite the good performance of the proposed baseline, the overall evaluations indicate that the problem of prominent object detection and recognition is a challenging task that is still worth investigating further.*

## 1. Introduction

The computer vision community has come a long way in detection and recognition of objets. Comparing the current achievements to the robot visual obstacle avoidance techniques in early 80s [13], they all have one thing in common, that is, they are inspired by the human vision system. In the annals of computer vision history, different techniques

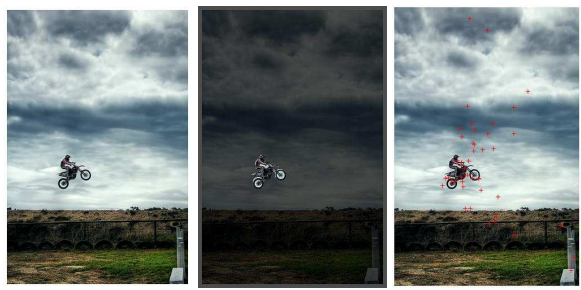

Machine-generated caption: It's graffiti on a cloudy day!

Figure 1. An example image, the most important object selected by human, fixation points, and a caption generated by a machine using a neural translation technique to motivate the necessity of knowing the key element in a scene.

and approaches have been explored for object detection and recognition. To date, we are capable of detecting and recognizing all the objects in a scene using semantic segmentation techniques [12]. With a high degree of accuracy, we can localize and recognize the objects of interest using efficient object localization and recognition methods [15, 16]. There is, however, one question to be answered in any scene understanding problem, particularly when reasoning is involved. That is, "What is the importance of each object in the scene?"

Moving towards answering this question, we first tackle an easier problem involving the key/prominent object of the scene. We here define the problem of prominent object detection and recognition. We, however, foresee extending the problem formulation to all the object instances in the scene.

**Why the prominent object is important?** Let us play a game, that is, given an image, make a story. To make a story from an image, intuitively speaking, we often identify the key subject and build our story around it by taking the other elements into account. For example in case of Figure 1, one may write, "John, who is a talented motor-racer, decided to spend the weekend..." That is, given the image, we realize the story should be around a person and his abilities as a motorcyclist. In other words, prior to any imagination and reasoning, we are using our ability to identify the prominent object.

In many high-level inferences that involve knowledge representation (KR) from images, identification of the key visual element can be of benefit. Examples of such inferences can be found in applications, including, image captioning [22], creative image caption generation [10], visual question answering [1], etc. It is worth noting that there exists end-to-end neural techniques for such applications, which should not be confused with the KR-based techniques. Nonetheless, such end-to-end techniques are not necessarily successful in key object identification as depicted in Figure 1, where the displayed caption is generated by a neural translation technique [19].

To address the question: "Where is the most important element/object in the scene?", the computer vision community has defined the task of saliency modeling, determining the locations a person pays attention to, and saliency detection, segmenting the most important object independent of the object type. Given the nature of the problem we are formulating in this paper, it is necessary to revisit these two questions under one umbrella. To achieve this end, we are solving the problem of finding, segmenting, and naming the most important object in the scene.

Problem formulation: *Given an image, determine the most influential item in the scene in terms of region of interest, pixel-level extent (segmentation), and object type.* In other words, we are looking forward for an algorithm that finds, segments and outputs the class label of the only one object that is the most influential item in the scene.

**How does the proposed problem relate to assitive vision and robotics?** The robot vision and active vision pipelines have been exploiting the concept of priority in perception by saliency-based pipelines for curtailing excess information and speeding up information processing. The same concept is also employed in obstacle aversion devices for visually impaired. We are, however, moving towards a new age of assistive techniques, relying on higher-level inferences, where a system can provide descriptions of the scene and gives recommendations. This necessitates us revisiting the existing pipelines, where the proposed problem formulation benefits the next generation of algorithms that rely on high-level inferences.

**Our contribution.** The main contribution of current manuscript is *revisiting the problem of saliency modeling and detection in conjunction with object recognition to identify and recognize the most prominent object in a scene*. We, thus, introduce a model that 1) finds the most important area in the image, 2) segments the most relevant object in that area, and 3) associates a class label with the segmented object. We evaluate the proposed method using a data, consisting of human fixations, object segmentations, and class labels.

## 2. Related studies

There are three main research areas that the current problem relates to. Each of these areas have provided solutions to one or two subtasks of the current problem. We briefly discuss them in this section.

**Saliency research.** The two aspects of saliency research, 1) saliency modeling, and 2) saliency detection, are associated with the problem of estimating the importance of image regions and segmenting important objects, respectively.

In saliency modeling, the human fixations are used as a measure of attention and importance. The saliency modeling is also recognized as fixation prediction. It can be translated to predicting a distribution that best matches human fixation density maps. The literature is replete with saliency models which going through is out of the scope of this paper. We thus refer the readers to review papers, e.g. [3].

On the other hand, saliency detection algorithms are dealing with segmenting the most important object. The algorithms in this class are often solving a classification problem to assign a binary label to pixels. The algorithms of saliency detection are well compared and detailed in [2], we thus avoid going through all of them here.

While part of the community may consider these two research areas different, they are highly related. The most relevant work to us is [11], which signifies this connection. They study the connection between fixations and segmentation with several saliency databases. Then, they propose a pipeline from fixations to salient objects by combining the techniques from the two aforementioned areas. Contrary to us, their method, akin to other saliency detection methods, does not produce any fixation density maps as a measurable output.

There is, however, a significant difference between the proposed baseline in this paper and the typical methods of saliency research. That is, the saliency problem formulation, which does not require the saliency methods to produce any class label, is different from ours.

**Object proposal and classification.** The research in proposing locations prone to contain objects and labeling those locations with object classes is a well studied and established domain. The main motivation behind object proposals is avoiding exhaustive search methods such as a sliding window in object detection pipelines. In the early days, the object proposal techniques were a separate pre-processing step. Recently, the tendency has been towards integrated pipelines by formulating the object proposal and detection as a multi-tasking problem, e.g., [6]. Despite the object proposal-based detectors are capable of providing series of areas of interest and their associated object class labels, they are not capable of discriminating the items based on their importance, i.e., all the detected objects are considered equal.

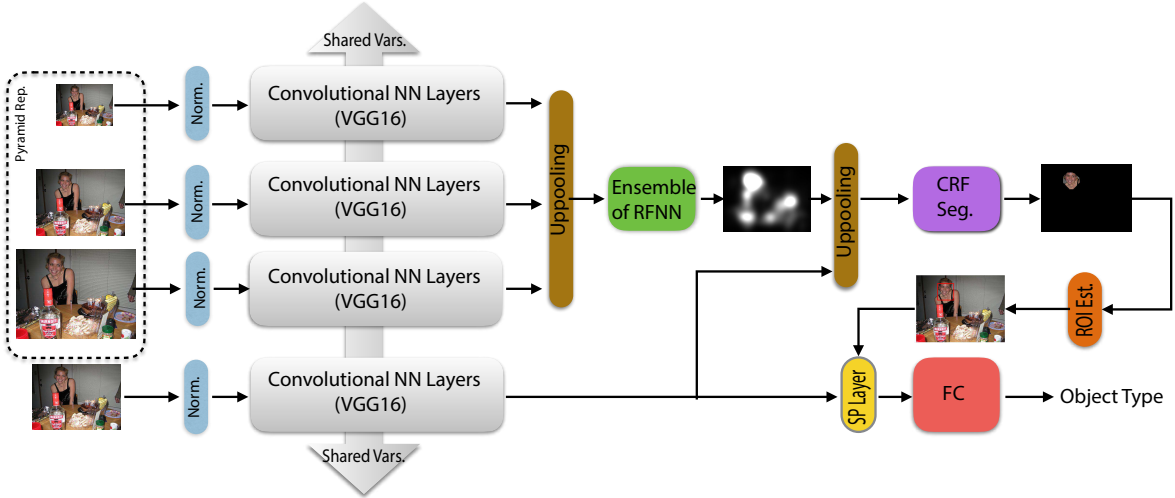

Figure 2. Overall overview of the proposed pipeline.

In object detection pipelines, an alternative approach to the object proposal methods is saliency-based techniques. These approaches, which are the connecting bridge between saliency research and object detection, initially compute a saliency map using a saliency modeling technique and perform a visual search starting from the maximally salient area, e.g., using a winner-take-all approach as described in [8]. These approaches have been favored for efficient object detection and localization in neuromorphic computing, e.g. [9]. The saliency-based search techniques are capable of providing object importance. Nonetheless, they are not capable of segmenting the object of interest. Consequently, similar to object proposals, to localize the objects, they need to evaluate several hypothetical bounding boxes and rely on some non-maximal suppression operations.

While the saliency-based object detection pipelines do not require segmentation, adding segmentation criterion will reduce the necessity for multiple bounding-box evaluations by providing one exact region. Our proposed baseline is thus different from the saliency-based object detection pipelines because it is extended with the segmentation criterion.

**Semantic segmentation.** Assigning each pixel of the image a class label is recognized as semantic segmentation. There exist numerous research papers which are addressing this problem, e.g. [12]. Going through the semantic segmentation methods in details is beyond the scope of this article. There is one major difference between semantic segmentation and the problem defined in this paper. That is, semantic segmentation does not consider the importance of the segmented objects. In other words, the semantic segmentation algorithms provide a pixel-level segmentation of labeled items/objects in a scene, but they are not able to differentiate between their importance.

Furthermore, there is a conceptual difference between our proposed method and semantic segmentation techniques. Most of the semantic segmentation algorithms initially obtain an object score and build the semantic segmentation around it, which mimics a top-down pipeline. We, however, build the proposed baseline algorithm around a bottom-up approach in which the important area is identified by saliency and the shape grouping takes place prior to object detection.

## 3. Method

Figure 2 depicts an overall overview of the proposed baseline model. The model follows a hierarchal vision pipeline. That is, the processed image is first decomposed into features via convolutional neural networks (CNNs). The features are used in a randomly weighted feedforward network (RFNN) to identify the most salient area in the image. The saliency distribution and features are then used to segment the extent of the salient object. The extent of the segmented object is used as a region of interest (ROI) for object recognition after that. In the rest of this section, we explain the details of this pipeline.

**Convolutional layers.** The pipeline consists of two streams of convolutional neural networks, which are sharing their weights. The convolutional layers follow the VGG16 [20] architecture. The input to the saliency prediction stream consists of images with various resolutions corresponding to an image pyramid of sizes $\{800 \times 600, 400 \times 300, 224 \times 224\}$. The input to the object detection stream is resized to the standard VGG16 input of $224 \times 224$.

**Saliency detection and segmentation.** The saliency detection pipeline relies on a fixation prediction pipeline in order to detect the most salient area of the input image and exploits a Conditional Random Field (CRF)-based segmentation scheme to detect the extent of the most salient object. In other words, we first perform a saliency modeling, aka fixation prediction, in order to find the most salient area. The information of the salient area is used in conjunction with the image features to segment the most salient object.

*Saliency modeling.* To identify the salient region, we employ a saliency modeling pipeline using randomly weighted feedforward neural networks. We, thus, define the saliency as $\mathbf{S} = p(y|\boldsymbol{x}, m)$, where $y$ corresponds to pixel-level saliency of an image, $\boldsymbol{x}$ and $m$ correspond to image features and locations, respectively. Under independence assumption, the saliency is boiled down to $\mathbf{S} = p(y|\boldsymbol{x})p(y|m)$. Building on top of the iSEEL [21] framework, we employ an ensemble of neural predictors. Hence,

$$p(y|\boldsymbol{x}) = \sum_j \max(0, \tanh(y_j|\mathrm{u}_j(\boldsymbol{x}))), \quad (1)$$

where $y_j$ is the prediction from the output of the $j$th neural saliency predictor in the ensemble and $\mathrm{u}_j(\cdot)$ is a mapping function. To estimate the saliency, a mapping based on Randomly-weighted Feedforward Neural Network (RFNN) [18] is used. In particular, we employ Extreme Learning Machines (ELM) [7]. That is, to train each ensemble, having a set of training sample pairs $\{(\boldsymbol{x}_n, y_n)\}_{n=1}^N \subset \mathbb{R}^k \times \mathbb{R}^m$, and the neural responses of an ensemble $y_{j,n}$ for a sample input $\boldsymbol{x}_n$ defined as,

$$y_{j,n} = \mathrm{u}_j(\boldsymbol{x}_n) = \sum_{i=1}^L \gamma_{j,i} \mathrm{f}(\boldsymbol{\omega}_{j,i} \cdot \boldsymbol{x}_n + b_{j,i}), \quad (2)$$

where for the $j$th ensemble, $L$ is the number of hidden neurons, $\mathrm{f}(\cdot)$ is a nonlinear activation function, $\gamma_{j,i}$ is the output weight, $\boldsymbol{\omega}_{j,i} \in \mathbb{R}^k$ is the input weight vector, and $b_{j,i}$ is the bias of the $i$th hidden node, we can estimate the output weight $\gamma_i$ through *Moore-Penrose psudoinverse* while the input weights and bias are chosen randomly. In the matrix form, this is expressed as $\boldsymbol{\Gamma} = \mathbf{H}^\dagger \mathbf{Y}$, where $\mathbf{H} \in \mathbb{R}^{N \times L}$ is the matrix form of the activation functions.

Motivated by the usefulness of spatial priors in saliency modeling, we define $p(y|m) \sim \mathcal{N}(\boldsymbol{\mu}, \boldsymbol{\Sigma})$, where $\boldsymbol{\mu}$ is mean and $\boldsymbol{\Sigma}$ is the diagonal covariance matrix, estimated from fitting a 2D-Gaussian distribution to the human eye fixation. For this purpose, we used the Toronto database [4].

*Salient object segmentation.* Once we have determined the most salient area in a given image, the extent of the present object is determined. To achieve this end, we employ a Conditional Random Field (CRF) using saliency map and features from CNNs. In other words, we are formulating a binary labeling problem as an energy minimization where saliency contributes to the unary term and neural features are used to measure pairwise similarities between the pixels.

Defining an image array $\mathbf{I}$ with $M$ pixels in terms of $M$ random variables, $I_1, \ldots, I_M$, we assign every pixel a label, $I_i \in \mathcal{L}$, such that $\mathcal{L} = \{0, 1\}$, where 0 indicates non-object and 1 indicates object. The labeling is achieved by minimizing:

$$\mathrm{E}(\mathbf{I}, \mathbf{S}, \mathbf{F}) = \sum_{i=1}^M \Phi(I_i, S_i) + \sum_{i,j \in \mathcal{E}_i} \Psi(I_i, I_j, F_i, F_j), \quad (3)$$

where $\Phi(\cdot)$ is a unary energy function and $\Psi(\cdot)$ is a pairwise energy function, $S_i$ corresponds to the saliency of the $i$th pixel obtained by up-pooling the saliency map of the image, $F_i$ is associated with the CNN responses of the $i$th pixel after a up-pooling layer, and $\mathcal{E}_i$ is the set of neighboring pixels of $i$. The unary term is then defined as

$$\Phi(I_i, S_i) = \delta_{I_i,1}(1 - S_i) + \delta_{I_i,0} S_i, \quad (4)$$

where $\delta_{\cdot,\cdot}$ is the Kronecker delta function. The pairwise energy function is

$$\Psi(I_i, I_j, F_i, F_j) = 1 - \delta_{I_i, I_j} e^{-\|F_i - F_j\|}, \quad (5)$$

and it is designed to penalize neighboring pixels with different labels. It also castigates the pixels that have the same labels with respect to the dissimilarity of their features. The higher the pixels' dissimilarity is, the higher the penalization is. Following the steps of CRF-based segmentation techniques, we employ the binary optimization method of Rother et al. [17] in order to infer the extent of the salient object in an eight-connected pixel array setup.

**Object detection.** In the proposed framework, the object detection pipeline employs spatial pyramid pooling, where after the convolutional layers a spatial pooling layer maps the responses of the region of interest into the fully-connected (FC) layers. In our implementation, we use the special case of pyramid pooling with one pyramid level in consistency with the Fast R-CNN [6] implementation. To determine the region of interest, we utilize the result from CRF-based segmentation and determine the most campact bounding box around the connected component. The information of this ROI is given to the spatial pooling layer for feeding the suitable feature vector to the fully-connected layers.

**Training.** The training of the proposed framework is performed in multiple stages. The training stages are 1) learning to detect objects, and 2) training to find the important areas.

*Training the object detector.* The object detector in our framework is similar to the Fast R-CNN network with no location regression. Thus, instead of training an object detector from scratch, we adopt the necessary layers from the pre-trained Fast R-CNN network. The weights from the VGG16 are then shared between the several network streams.

*Training the saliency predictor.* The saliency prediction is carried out by an ensemble of randomly-weighted feed-forward network. Thus, having the weights of the VGG16 networks fixed, we learn the output weights of each neural unit in the ensemble from an eye fixation database. Please consult iSEEL model [21] for more details. The segmentation is formulated as a semi-supervised CRF energy minimization that relies on the saliency prediction.

**Why not to fine-tune the whole pipeline together?** While in theory it is possible to formulate the above pipeline as a multi-task problem, there exists one practical difficulty. That is, the data for the training of the pipeline as a multi-task problem is small. As will be discussed later, we have only 850 images suitable for evaluating the defined problem. It is also worth noting that we are introducing a simple and effective baseline method.

## 4. Experiments

**Data.** Evaluating a saliency-based model that predicts the most salient region of interest, segments the associated object, and assigns a class label to it, requires data that provides priority of objects in semantically segmented images.

To the best of our knowledge, the best suitable existing data is a subset of the SSOS [11] dataset, consisting of 850 images of the PASCAL VOC 2010 [5] dataset. This subset provides eye fixations from 8 subjects in a free-viewing task, where each image has been presented for 2 seconds with recalibration between every 25 images. Each image has full contextual segmentation for all the objects and items in the scene. The most salient object is also identified via a mouse click experiment by 12 independent subjects. For an object, the saliency value is defined by dividing the total number of received clicks to the number of subjects. The total number of segmented object categories is 459 including the 20 object class categories of PASCAL VOC 2010.

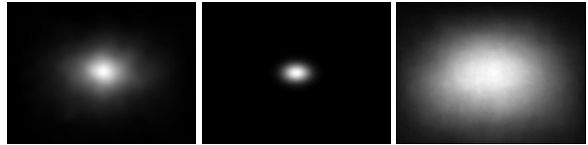

Figure 3. Center-bias. Left: Mean Eye Position (MEP), Middle: the normal distribution fitted to the fixation positions of the observers, used as chance model for fixation prediction, Right: Average Annotation Map (AAM) from segmentation ground truth.

**Evaluation protocol.** Our problem formulation requires us to find important regions, segment the associated objects in those regions, segment and classify them. The data used in the evaluation provides a ground truth for importance in terms of human fixations, the segmentation of objects and their class labels. Thus, to evaluate how well a model finds important areas, we rely on saliency modeling metrics. We chose the Normalized Scanpath Score (NSS) [14] to measure the consistency between human and machine. Having the set of observers' fixation locations, $\{(x_i^f, y_i^f)\}_{i=1}^{N}$, and a saliency map $\mathbf{S}$, the NSS is defined as,

$$NSS = \frac{1}{N} \sum_{i=1}^{N} \frac{\mathbf{S}(x_i^f, y_i^f) - \mu_{\mathbf{S}}}{\sigma_{\mathbf{S}}}, \qquad (6)$$

where $\mu_{\mathbf{S}}$ is the mean and $\sigma_{\mathbf{S}}$ is the standard deviation of the saliency map.

The quality of segmentation is evaluated in terms of the $F_\beta$-measure, which is a weighted harmonic mean with a non-negative weight, $\beta$. The $F_\beta$-measure is a metric that summarizes the traditional precision and recall metrics into one as follows:

$$F_\beta = \frac{(1+\beta^2) Precision \cdot Recall}{\beta^2 Precision + Recall}. \qquad (7)$$

Following the saliency segmentation techniques and surveys [2], we use $\beta^2 = 0.3$. We will also include the precision and recall values in order to understand the model's shortcoming better.

The evaluation of detected object class categories is carried out in terms of accuracy, $ACC = (TP + TN)/N$, where $N$ is the total number of object instances, $TP$ and $TN$ are the true positive and true negative detections, respectively.

It is worth noting that the object proposal and detection algorithms rely on mean precision because they are retrieving as many instances of an object class as possible. We are, however, interested in the most important object. The accuracy metric is, thus, sufficient in the current problem setup.

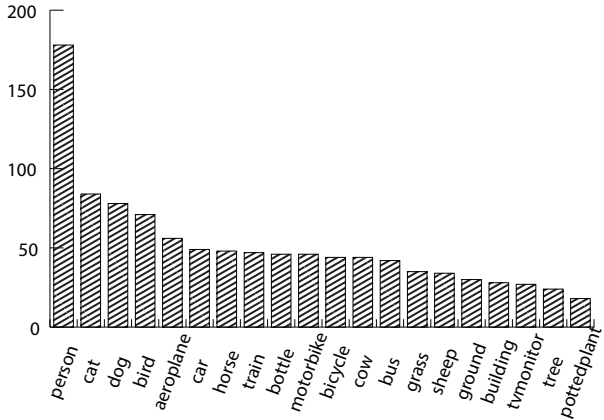

Figure 4. Object instances: the number of top referenced most important object instances in the image set.

We further complement accuracy with confusion matrices for gaining a better understanding of the performance of the pipeline.

**Analysis of data and chance level.** To obtain an insight about the underlying statistics of this database, we analyzed some of the properties of the existing data in order to establish a chance level for the evaluation protocol. We computed the mean eye position of the observers. We furthermore fitted a 2D normal distribution, $\mathcal{N}(\boldsymbol{\mu}, \boldsymbol{\Sigma})$, to the normalized fixation points of all the observers, where the mean is $\boldsymbol{\mu} = (0.0119, 0.0324)$ and the diagonals of the covariance matrix $\boldsymbol{\Sigma}$ are $\sigma_1 = 0.09$ and $\sigma_1 = 0.1$, respectively. As indicated in Figure 3, there exists a strong center bias in the data. It is thus not surprising that evaluating the estimated 2D normal distribution against users' fixations produces an NSS value 1.432. We chose this performance as the chance level performance for the NSS score.

We conducted a similar analysis using the Average Annotation Map (AAM), akin to [2], in order to build a chance model for the segmentation task. As depicted in Figure 3, there is a high tendency for having an object positioned in the center of the image. The performance of a chance model for segmentation task is $F_\beta = 0.36$.

We also analyzed the distribution of object category instances among the most important objects in the images. The results are summarized in Figure 4. From 459 object classes, instances of only 25 of the classes are identified as the most important object, with class category "person" having been chosen 178 times as the most influential object in the whole image set. Thus, the chance classifier for the database performs at $ACC = 0.21$.

| Model | $NSS$ |
|---|---|
| Human upper-bound | 3.09 |
| Proposed model | 2.05 |
| Chance | 1.43 |

Table 1. The performance of baseline in predicting the region of interests in terms of consistency with human eye fixations.

| Model | $F_\beta$ | $Prec.$ | $Rec.$ |
|---|---|---|---|
| Proposed model | 0.51 | 0.68 | 0.41 |
| Chance | 0.36 | 0.35 | 0.58 |

Table 2. The assessment of the segmentations produced for the most important object in the scene. $F_\beta$, precision ($Prec.$), and recall ($Rec.$), are reported.

| Model | $ACC$ |
|---|---|
| Proposed model | 0.83 |
| Chance | 0.21 |

Table 3. The overall accuracy of object detection.

## 5. Results

Table 1 summarizes the performance of the proposed model in predicting the correct region of interest in terms of NSS. The proposed model performs clearly better than chance. There is, however, a significant gap between the proposed model and the human upper-bound performance.

Table 2 reports the $F_\beta$ performance for the segmentations generated by our model. While the performance of the proposed method is above chance, there exists room for improvements in this task. Having a closer look into the mean value of precision and recall reveals that in comparison to the chance model on the average 1) the proposed segmentation has fewer false positive, and 2) the proposed model under-segments the objects. Overall the precision and recall values indicate that the accuracy of the segmentation pipeline has to be improved in order to reduce the number of misses and increase the true positive rate, while preventing false positive detections.

The third performance criterion to check is the accuracy of the object class category detection, summarized in Table 3. We further complemented this results with the confusion matrix for more detailed results. Figure 5 presents the confusion matrix. As depicted, several object class categories are identified well. Nonetheless, a small number of class categories are not recognized. The reason is that we are relying on Fast R-CNN detector capable of detecting only the 20 object categories of PASCAL VOC. On the other hand, the prominent objects necessarily do not belong to this subset 20 object classes even though belong to a larger subset of the 459 contextually annotated items.

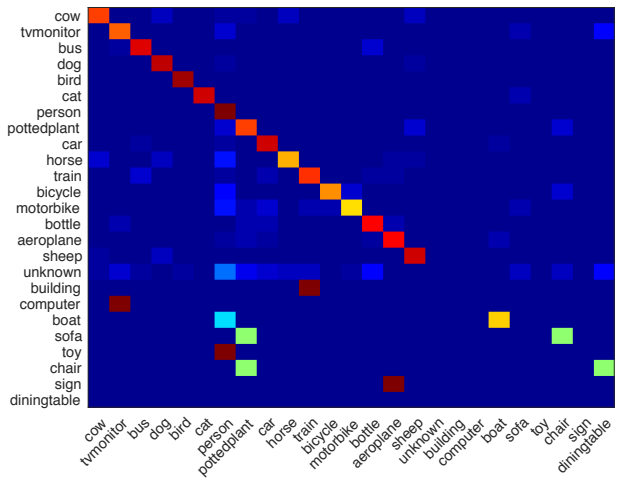

Figure 5. Confusion matrix of detection.

**Visual examples.** Figure 6 visualizes several examples of the output of the proposed model and the ground truth data. We chose a variety of scenarios where the method works well or breaks. In general, the segmentation algorithm in the proposed framework produces false positive areas affecting the overall performance of the algorithm.

The second noticeable problem is predicting the importance of the objects mistakenly. This often happens when the saliency prediction algorithm fails. For instance, in the fourth row of Figure 6, we observe that despite the object is associated with a correct label, it is not the most prominent object of the scene. Comparing the fixation prediction maps to the ground truth, we can even observe the slight difference in the importance of the objects visually. In case of the third row, the method is incapable of differentiating the importance of buses.

The incorrect region of interest detection and segmentation can affect the object detection result. This phenomena is evident in the fifth and sixth rows of Figure 6. Eventually, the object detector may make a mistake in associating the correct class label to an item as we can observe in the last two rows.

## 6. Discussion

In this paper, we revisited the problem of saliency and object recognition to formulate the research problem of the most prominent object detection and recognition. We proposed a simple deep learning-based method that can be used as a baseline for further research. The performance of the proposed method was evaluated in terms of consistency with human fixations, most important object segmentation, and the detected object class label.

**Do we need a new problem formulation?** The three domains of (1) saliency modeling, (2) saliency detection, and (3) object recognition are saturated by various algorithms. Most of the recent state-of-the-art algorithms for each problem are achieving close to human performance in each of these tasks. From a bio-inspired vision system perspective, we are, however, still far from emulating a primate vision system and its full cognitive powers. Just thinking about how each of us detect and recognize the objects in our surrounding environments, is a good motive to revisit all of these individual problems under one umbrella.

Furthermore, as already discussed in the introduction section, the nature of KR-based vision solutions benefits from the information that can be obtained by solving the prominent object detection and recognition problem first.

**Does segmentation help?** Advocates of saliency-based object search techniques may argue that the segmentation is not necessary. This may be partially true depending on the purpose of the object detection. The requirement for segmentation becomes, however, a necessity when the object detection is an input to an inference pipeline that requires reasoning about the relation of objects and their pixel-wise extent as in semantic segmentation applications.

**Are 850 images enough?** We are using 850 images as a test set. The training data can, however, be obtained from other sources like the PASCAL (for object detection), MIT1003 (for saliency modeling), and MSRA10k (for saliency segmentation). Thus, while the 850 images are enough to test an algorithm, the lack of data to cover all the three ground-truth types with the same input stimuli makes solving the problem a challenge for fully supervised methods. This further leaves room for fair evaluation of semi-supervised methods in conjunction with supervised methods and gives us a better understanding about the generalization power of the supervised methods.

**What about time to detect and efficiency?** Time to detect and efficiency (in terms of the number of computations and the used energy) are other important factors to measure. They indicate how easily a method can be employed in a real-time scenario. In this manuscript, we are, however, focusing on the problem formulation and introduction rather than benchmarking methods and techniques for solving it. We, thus, leave these questions to be addressed later.

**Future research direction.** We are expecting the future research direction should focus on extending the data towards priority of all the instances of objects in the scene and scaling up the amount of data.

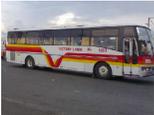

Figure 6. Visual examples: Rows 1-2: correct detections. Rows 3-4: correct detection, but incorrect most important object. Rows 5-6: wrong important region of interest and wrong detection. Rows: 7-8: correct region of interest, incorrect object detection.

## 7. Conclusions

In this paper, we introduced the problem of *prominent object detection and recognition* and defined a baseline algorithm for it. The challenge is in finding the most important region of interest, segmenting the relevant item/object in that area, and assigning it an object class label.

The proposed method follows a feed-forward bio-inspired vision system pipeline using deep-learning-based features as the main ingredient. The pipeline is modular and trained in multiple stages. We avoid fine-tuning on the evaluation dataset. The output of the pipeline includes: a predicted fixation density map, the most important item's segment, and the detected class label. Thus, the method's output is evaluated against three types of ground-truth data, including, human fixation points, the human annotated segmentation masks, and object labels.

Despite the good performance of the proposed baseline, the overall evaluations indicate that the proposed problem is a challenging one. Solving this challenging problem can 1) help us getting closer to emulating primate vision systems, and 2) benefit the visual inference pipelines with KR-based engines.

## References

[1] S. Antol, A. Agrawal, J. Lu, M. Mitchell, D. Batra, C. L. Zitnick, and D. Parikh. VQA: Visual Question Answering. In *International Conference on Computer Vision (ICCV)*, 2015. 2

[2] A. Borji, M.-M. Cheng, H. Jiang, and J. Li. Salient object detection: A benchmark. *IEEE TIP*, 24(12):5706–5722, 2015. 2, 5, 6

[3] A. Borji and L. Itti. State-of-the-art in visual attention modeling. *IEEE Transactions on Pattern Analysis and Machine Intelligence*, 99(PrePrints), 2012. 2

[4] N. D. B. Bruce and J. K. Tsotsos. Saliency based on information maximization. In Y. Weiss, B. Schölkopf, and J. Platt, editors, *Advances in Neural Information Processing Systems 18*, pages 155–162, MIT Press, 2006. MIT Press. 4

[5] M. Everingham, L. Van Gool, C. K. I. Williams, J. Winn, and A. Zisserman. The PASCAL Visual Object Classes Challenge 2010 (VOC2010) Results. http://www.pascal-network.org/challenges/VOC/voc2010/workshop/index.html. 5

[6] R. Girshick. Fast r-cnn. In *2015 IEEE International Conference on Computer Vision (ICCV)*, pages 1440–1448, Dec 2015. 3, 5

[7] G.-B. Huang, Q.-Y. Zhu, and C.-K. Siew. Extreme learning machine: a new learning scheme of feedforward neural networks. In *2004 IEEE International Joint Conference on Neural Networks (IEEE Cat. No.04CH37541)*, volume 2, pages 985–990 vol.2, July 2004. 4

[8] L. Itti, C. Koch, and E. Niebur. A model of saliency-based visual attention for rapid scene analysis. *Pattern Analysis and Machine Intelligence, IEEE Transactions on*, 20(11):1254 – 1259, nov. 1998. 3

[9] R. Kasturi, D. B. Goldgof, R. Ekambaram, G. Pratt, E. Krotkov, D. D. Hackett, Y. Ran, Q. Zheng, R. Sharma, M. Anderson, M. Peot, M. Aguilar, D. Khosla, Y. Chen, K. Kim, L. Elazary, R. C. Voorhies, D. F. Parks, and L. Itti. Performance evaluation of neuromorphic-vision object recognition algorithms. In *2014 22nd International Conference on Pattern Recognition*, pages 2401–2406, Aug 2014. 3

[10] S. Li, G. Kulkarni, T. L. Berg, A. C. Berg, and Y. Choi. Composing simple image descriptions using web-scale n-grams. In *Proceedings of the Fifteenth Conference on Computational Natural Language Learning*, CoNLL '11, pages 220–228, Stroudsburg, PA, USA, 2011. Association for Computational Linguistics. 2

[11] Y. Li, X. Hou, C. Koch, J. M. Rehg, and A. L. Yuille. The secrets of salient object segmentation. In *2014 IEEE Conference on Computer Vision and Pattern Recognition*, pages 280–287, June 2014. 2, 5

[12] J. Long, E. Shelhamer, and T. Darrell. Fully convolutional networks for semantic segmentation. In *2015 IEEE Conference on Computer Vision and Pattern Recognition (CVPR)*, pages 3431–3440, June 2015. 1, 3

[13] H. P. Moravec. Rover visual obstacle avoidance. In *Proceedings of the 7th International Joint Conference on Artificial Intelligence - Volume 2*, IJCAI'81, pages 785–790, San Francisco, CA, USA, 1981. Morgan Kaufmann Publishers Inc. 1

[14] R. J. Peters, A. Iyer, L. Itti, and C. Koch. Components of bottom-up gaze allocation in natural images. *Vision Research*, 45:2397–2416, 2005. 5

[15] J. Redmon, S. Divvala, R. Girshick, and A. Farhadi. You only look once: Unified, real-time object detection. In *The IEEE Conference on Computer Vision and Pattern Recognition (CVPR)*, June 2016. 1

[16] S. Ren, K. He, R. Girshick, and J. Sun. Faster R-CNN: Towards real-time object detection with region proposal networks. In *Neural Information Processing Systems (NIPS)*, 2015. 1

[17] C. Rother, V. Kolmogorov, V. Lempitsky, and M. Szummer. Optimizing binary mrfs via extended roof duality. In *Proc Comp. Vision Pattern Recogn. (CVPR)*, June 2007. 4

[18] W. F. Schmidt, M. A. Kraaijveld, and R. P. W. Duin. Feedforward neural networks with random weights. In *Proceedings., 11th IAPR International Conference on Pattern Recognition. Vol.II. Conference B: Pattern Recognition Methodology and Systems*, pages 1–4, Aug 1992. 4

[19] R. Shetty, H. R.-Tavakoli, and J. Laaksonen. Exploiting scene context for image captioning. In *Proceedings of the 2016 ACM Workshop on Vision and Language Integration Meets Multimedia Fusion*, iV&L-MM '16, pages 1–8, New York, NY, USA, 2016. ACM. 2

[20] K. Simonyan and A. Zisserman. Very deep convolutional networks for large-scale image recognition. In *ICLR*, 2015. 4

[21] H. R. Tavakoli, A. Borji, J. Laaksonen, and E. Rahtu. Exploiting inter-image similarity and ensemble of extreme learners for fixation prediction using deep features. *Neurocomputing*, pages –, 2017. 4, 5

[22] O. Vinyals, A. Toshev, S. Bengio, and D. Erhan. Show and tell: Lessons learned from the 2015 mscoco image captioning challenge. *IEEE Transactions on Pattern Analysis and Machine Intelligence*, 39(4):652–663, April 2017. 2